\documentclass{article}
\usepackage{spconf,amsmath,amssymb,graphicx,hyperref}
\usepackage{booktabs,multirow,tabularx,balance}
\usepackage[table]{xcolor}
\usepackage{enumitem}
\usepackage{cite}

\hypersetup{hidelinks}

\newcommand{\benchmark}{\textsc{ScreenASR-Bench}}

\title{ScreenASR-Bench: A Tiered Benchmark for Speech Recognition with Visual Context}
\name{Mingwei Fu$^{1}$, Zheshu Song$^{2}$, Zhikai Zhou$^{2}$,
Shen Huang$^{2}$, Shidong Shang$^{2}$, Mengxiao Zhu$^{1,*}$
\thanks{$^{*}$Corresponding author.}}
\address{$^{1}$University of Science and Technology of China \\
$^{2}$Tenlai Team, China}

\begin{document}
\ninept
\maketitle

\begin{abstract}
Current ASR models perform well on general benchmarks but still struggle with acoustically ambiguous words. Visual context can aid recognition in screen-based scenarios, yet existing benchmarks provide limited insight into the benefits of different types of visual evidence.
To address this gap, we introduce \benchmark , a tiered benchmark with three progressively challenging levels: Explicit Textual Evidence (ETE), based on direct on-screen text; Explicit Visual Evidence (EVE), using visible content rather than on-screen text; and Implicit Semantic Evidence (ISE), requiring scene-level inference. 
Extensive experiments on ASR models, Large Audio-Language Models (LALMs) and Omni models show that incorporating visual context substantially improves recall over the audio-only setting across all three model families. The gains are largest for ETE and smallest for ISE, indicating that models leverage direct visual evidence more effectively than indirect semantic cues.
To facilitate future research on visually grounded speech recognition, we publicly release \benchmark.\footnote{\url{https://huggingface.co/datasets/MingweiFu/ScreenASR-Bench}}
\end{abstract}


\section{Introduction}
\label{sec:intro}

Recent advances in speech foundation models~\cite{review,salmonn,slam,radford2023whisper,qwen2-audio,index-asr,funaudiollm} and large-scale datasets have driven substantial progress in automatic speech recognition (ASR), yielding low word error rate (WER) on standard benchmarks~\cite{panayotov2015librispeech,aishell1,aishell2,MLS,fleurs}. However, some ambiguous keywords remain difficult to recognize from speech alone, even when most of the surrounding utterance is transcribed correctly. Such localized errors may contribute little to aggregate WER while substantially changing the intended meaning. 
In audio-visual settings, on-screen text, visible entities or actions, and scene semantics can provide contextual visual evidence for recognizing these challenging words. This raises a key question: \textbf{how effectively can current speech models exploit different levels of visual information for ambiguous keywords recognition?}

Prior studies have shown that visual context can improve recognition when lexical information is difficult to recover from the acoustic signal. Ghorbani et al.~\cite{ghorbani2021listen} demonstrate improved masked-word recovery with visual context, while Gabeur et al.~\cite{gabeur2022avatar} report larger visual gains under challenging acoustic conditions. Chen et al.~\cite{chen2026ocr} use on-screen text to enhance ASR, and Zhao et al.~\cite{zhao2025chineselips} combine lip-reading with textual and graphical information from presentation slides. More recently, Tian et al.~\cite{tian2026seeing} extend the scope of visual context to richer and more diverse scenes, including scenarios that require multimodal reasoning to resolve acoustic ambiguities. These studies demonstrate the value of visual information for speech recognition, but they generally treat visual context as a whole, without distinguishing between visual cues that differ in how directly they indicate the target keyword.

A similar limitation persists in existing audio-visual ASR benchmarks.
SlideSpeech~\cite{wang2024slidespeech} provides a large-scale slide-enriched corpus, while SlideAVSR~\cite{wang2024slideavsr} introduces a benchmark for recognizing technical terminology with presentation slides. Sinhamahapatra et al.~\cite{sinhamahapatra2025slides} similarly evaluate multimodal speech recognition in conference presentations, and VAPO~\cite{hu2026vapo} introduces SlideASR-Bench for entity-rich slide-enhanced recognition. These benchmarks provide valuable resources for evaluating visually assisted ASR, but they
treat all cases equally and do not distinguish different levels of visual evidence. As a result, it remains unclear which type of visual evidence a model can actually exploit.

\begin{figure}[t]
    \centering
    \includegraphics[width=1\columnwidth]{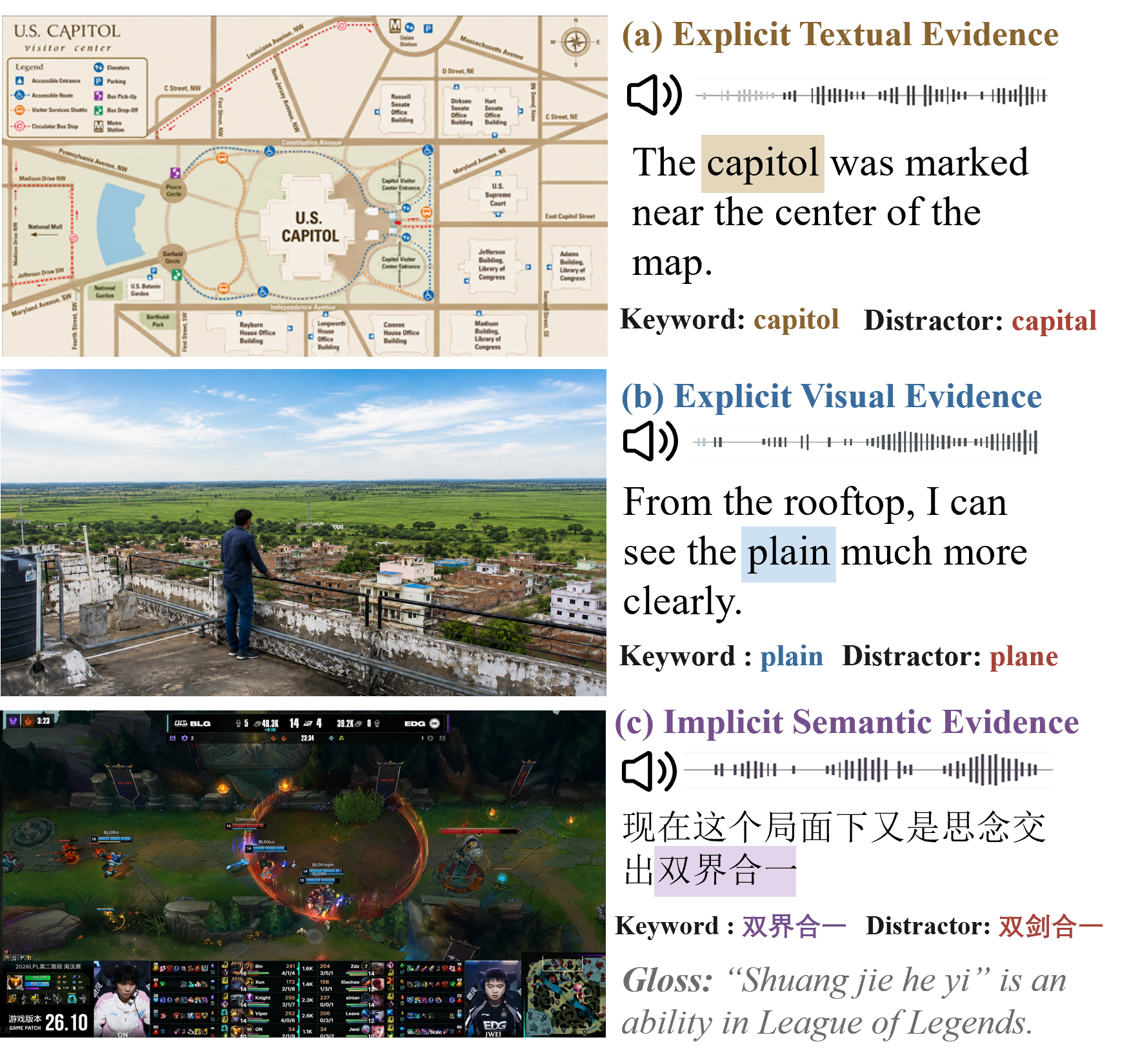}
    \par\vspace{-7pt}
    \caption{Representative examples of the three visual-evidence levels in \benchmark. From ETE to ISE, the semantic distance between the target keyword and its visual evidence increases.}
    \label{fig:taxonomy}
    \vspace{-3pt}
\end{figure}

\begin{figure*}[t]
    \centering
    \includegraphics[width=1\textwidth]{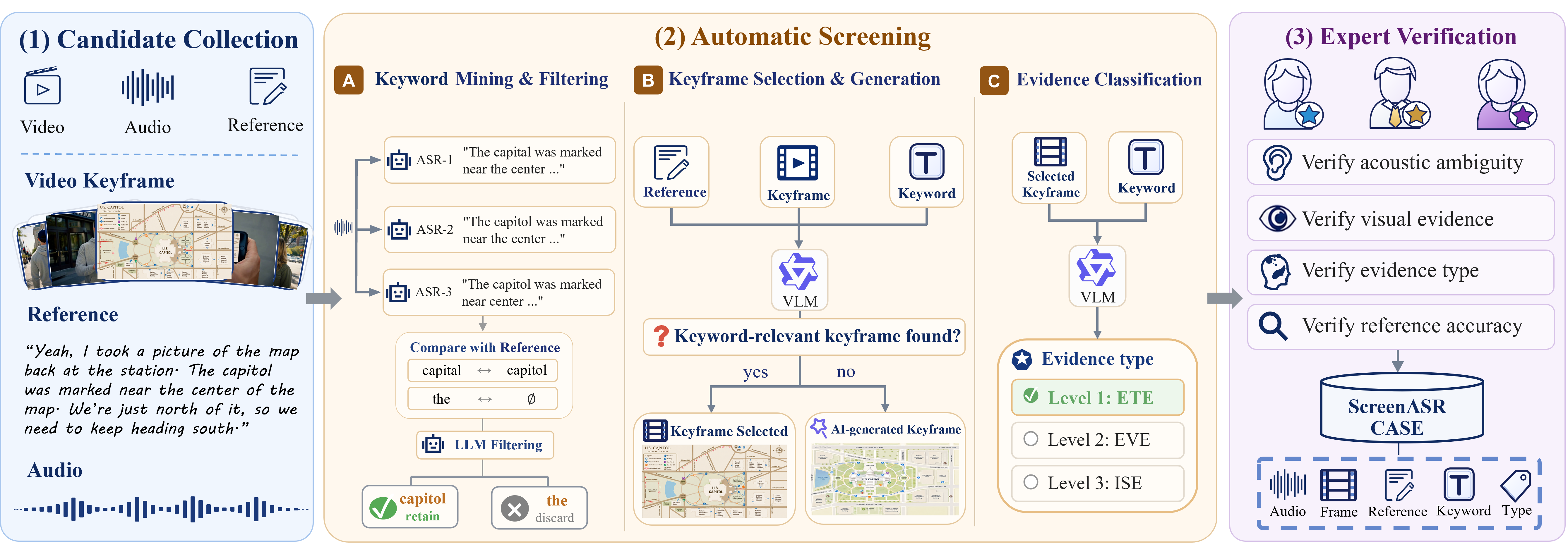}
    \vspace{-0.4em}
    \caption{Overview of the \benchmark\ construction pipeline, consisting of three stages: (1) candidate collection, (2) automatic screening, and (3) expert verification. Candidate keywords are mined from multi-ASR results and the VLM refers to Qwen3.5-35B-A3B.}
    \label{fig:pipeline}
\end{figure*}

To address this gap, we present \benchmark , a tiered benchmark for evaluating how models exploit different forms of visual evidence. It contains 2,002 cases collected from publicly accessible online videos and each case consists of a speech segment, a corresponding keyframe, a reference transcription, and a designated target keyword. As illustrated in Figure~\ref{fig:taxonomy}, cases are categorized into three levels according to the relationship between the target keyword and the visual evidence: Explicit Textual Evidence (ETE), Explicit Visual Evidence (EVE), and Implicit Semantic Evidence (ISE), corresponding respectively to readable text, directly visible referents that are not shown as text, and indirect scene-level semantics. From ETE to ISE, the relationship between the target keyword and its visual evidence becomes progressively less direct, increasing the difficulty of effectively leveraging the visual context. 
In summary, the contributions of this paper are as follows:

\begin{itemize}[leftmargin=0.5cm]
    \item We present \benchmark, a tiered benchmark for evaluating how effectively speech models exploit different levels of visual evidence for challenging lexical recognition.

    \item ASR models, LALMs, and Omni models are systematically evaluated on \benchmark, providing a comprehensive analysis of their ability to leverage visual context across the three evidence categories.

    \item We publicly release \benchmark\ to facilitate reproducible research and promote further advances in multimodal ASR.
\end{itemize}

\section{\benchmark}
\label{sec:benchmark}

In this section, we describe the construction pipeline of \benchmark, define the three levels
of visual evidence and analyze the statistics of the benchmark.




\begin{table*}[!t]
\centering
\caption{Model-level WER and Recall. Vision denotes the visual input: -- for audio-only, Caption for a VLM-generated keyframe caption, and Image for direct keyframe input. ETE, EVE, and ISE are the three evidence levels. Best and second-best results are bold and underlined.}
\label{tab:model_results}
\small
\setlength{\tabcolsep}{4pt}
\renewcommand{\arraystretch}{1.03}
\newcommand{\tblhead}[1]{{\fontsize{10}{11}\selectfont\bfseries #1}}
\newcommand{\tblsubhead}[1]{{\fontsize{9}{10}\selectfont\bfseries #1}}
\begin{tabularx}{\textwidth}{l@{\hspace{8pt}}c*{8}{>{\centering\arraybackslash}X}}
\toprule
\multirow{2}{*}{\tblhead{Model}} &
\multirow{2}{*}{\tblhead{Vision}} &
\multicolumn{2}{c}{\tblhead{ETE}} &
\multicolumn{2}{c}{\tblhead{EVE}} &
\multicolumn{2}{c}{\tblhead{ISE}} &
\multicolumn{2}{c}{\tblhead{Overall}} \\
\cmidrule(lr){3-4}\cmidrule(lr){5-6}\cmidrule(lr){7-8}\cmidrule(lr){9-10}
& & \tblsubhead{WER} & \tblsubhead{Recall} & \tblsubhead{WER} & \tblsubhead{Recall} & \tblsubhead{WER} & \tblsubhead{Recall} & \tblsubhead{WER} & \tblsubhead{Recall} \\

\specialrule{\lightrulewidth}{0pt}{0pt}
\rowcolor{gray!18}
\multicolumn{10}{c}{\textbf{ASR Models}} \\
\specialrule{\lightrulewidth}{0pt}{0pt}
FireRedASR2-LLM & -
& 7.02 & 35.11 & 5.34 & 43.07 & 6.15 & 34.62 & 6.22 & 37.87 \\
\addlinespace[1pt]
\rowcolor{gray!12}
MiMo-V2.5-ASR & -
& 6.97 & 47.07 & 4.80 & 59.28 & 5.22 & 54.99 & 5.77 & 53.43 \\
\addlinespace[1pt]
\multirow{2}{*}{Qwen3-ASR-1.7B} & -
& 7.88 & 34.10 & 5.99 & 42.52 & 6.47 & 39.10 & 6.87 & 38.37 \\
& Caption
& 6.08 & 77.86 & 5.34 & 52.77 & 6.03 & 44.40 & 5.81 & 60.58 \\
\addlinespace[1pt]
\rowcolor{gray!12}
& -
& 7.43 & 45.55 & 5.37 & 58.45 & 6.00 & 53.16 & 6.36 & 52.08 \\
\rowcolor{gray!12}
\multirow{-2}{*}{Qwen-Audio-3.0-ASR} & Caption
& 6.80 & 47.07 & 4.81 & 60.53 & 5.57 & 54.18 & 5.80 & 53.68 \\
\addlinespace[1pt]
\multirow{2}{*}{Seed-ASR-2.0} & -
& 5.68 & 48.28 & 4.01 & 62.22 & 4.57 & 51.43 & 4.82 & 54.09 \\
& Caption
& {4.19} & 82.57 & \underline{3.50} & 70.74 & {4.20} & 58.28 & {3.96} & 72.34 \\

\specialrule{\lightrulewidth}{0pt}{0pt}
\rowcolor{gray!18}
\multicolumn{10}{c}{\textbf{Large Audio-Language Models}} \\
\specialrule{\lightrulewidth}{0pt}{0pt}
\multirow{2}{*}{MOSS-Audio-8B-Instruct} & -
& 7.19 & 40.71 & 5.07 & 51.11 & 9.32 & 46.44 & 7.01 & 45.87 \\
& Caption
& 5.20 & 81.42 & 4.75 & 58.59 & 5.59 & 50.10 & 5.15 & 65.48 \\
\addlinespace[1pt]
\rowcolor{gray!12}
& -
& 9.04 & 43.00 & 6.84 & 51.66 & 14.28 & 45.42 & 9.63 & 46.72 \\
\rowcolor{gray!12}
\multirow{-2}{*}{MiMo-Audio-7B-Instruct} & Caption
& 6.90 & 82.06 & 6.04 & 60.25 & 6.56 & 51.12 & 6.52 & 66.58 \\
\addlinespace[1pt]
\multirow{2}{*}{StepAudio-2.5-Chat} & -
& 9.39 & 45.93 & 6.57 & 52.77 & 7.51 & 48.88 & 7.94 & 49.12 \\
& Caption
& 9.46 & 71.25 & 6.95 & 56.09 & 8.25 & 51.53 & 8.29 & 60.93 \\

\specialrule{\lightrulewidth}{0pt}{0pt}
\rowcolor{gray!18}
\multicolumn{10}{c}{\rule[-0.35em]{0pt}{1.35em}\textbf{Omni Models}} \\
\specialrule{\lightrulewidth}{0pt}{0pt}
\multirow{3}{*}{Qwen3-Omni} & -
& 6.81 & 56.23 & 4.69 & 66.62 & 5.32 & 59.47 & 5.70 & 60.78 \\
& Image
& 5.85 & 80.41 & 4.26 & 75.90 & 6.57 & 64.56 & 5.49 & 74.89 \\
& Caption
& 4.47 & 79.13 & \underline{3.50} & 73.68 & \underline{3.98} & 66.60 & 4.01 & 74.09 \\
\addlinespace[1pt]
\rowcolor{gray!12}
& -
& 11.39 & 62.98 & 5.69 & 68.42 & 10.68 & 63.95 & 9.25 & 65.18 \\
\rowcolor{gray!12}
& Image
& 6.72 & \underline{88.68} & 8.32 & 77.01 & 6.10 & 71.28 & 7.11 & \textbf{80.19} \\
\rowcolor{gray!12}
\multirow{-3}{*}{MiMo-V2.5} & Caption
& 5.35 & \textbf{90.20} & 4.27 & 74.93 & 4.79 & 69.86 & 4.83 & \underline{79.69} \\
\addlinespace[1pt]
\multirow{3}{*}{Qwen3.5-Omni-Plus} & -
& 5.97 & 62.09 & 4.24 & 69.81 & 4.98 & 65.99 & 5.12 & 65.83 \\
& Image
& 5.10 & 81.81 & 3.51 & \textbf{78.09} & 4.50 & \textbf{73.12} & 4.40 & 78.33 \\
& Caption
& \textbf{3.68} & 84.86 & \textbf{3.00} & \underline{77.95} & \textbf{3.76} & \underline{72.51} & \textbf{3.47} & 79.33 \\
\addlinespace[1pt]
\rowcolor{gray!12}
& -
& 6.85 & 53.56 & 4.81 & 58.31 & 5.44 & 57.03 & 5.79 & 56.13 \\
\rowcolor{gray!12}
& Image
& 5.23 & 84.27 & 4.09 & 71.33 & 4.65 & 66.80 & 4.69 & 75.29 \\
\rowcolor{gray!12}
\multirow{-3}{*}{Doubao-Seed-2.0-lite} & Caption
& \underline{4.17} & 85.68 & \underline{3.50} & 72.58 & 4.19 & 66.19 & \underline{3.94} & 76.14 \\
\bottomrule
\end{tabularx}
\end{table*}

\subsection{Dataset Construction Pipeline}
\label{sec:construction}

Constructing \benchmark\ requires both sufficient scale for reliable evaluation and sufficient annotation accuracy for use as ground truth; manual annotation alone does not scale, while fully automatic construction does not ensure label reliability. Therefore, we adopt a three-stage pipeline combining automatic screening with expert verification, as shown in Figure~\ref{fig:pipeline}.

\noindent\textbf{Stage 1: Candidate Collection.}
\benchmark\ is built from publicly available online videos.
Each candidate instance originates from a video segment and consists of an audio segment, its reference transcription, and a set of candidate keyframes. The keyframes are sampled from the video according to the timestamps of the audio segment. Since some of the reference transcriptions are provided in inverse text normalization (ITN) form, we apply an LLM-based text normalization (TN) step to convert all references into a unified textual format.

\noindent\textbf{Stage 2: Automatic Screening.}
This stage converts raw candidates into keyword--keyframe pairs with preliminary evidence labels, and proceeds in three steps.

\textit{(i) Keyword mining and filtering.}
We transcribe each audio segment with multiple ASR systems and align their hypotheses with the normalized reference; the words or phrases where the hypotheses disagree with the reference are taken as candidate keywords.
Using multiple recognizers instead of a single one prevents the benchmark from being biased toward the error patterns of any specific model, so that the mined keywords are those that are
difficult for ASR systems in general rather than for one particular system. Since many mismatches involve filler words or common function words, we further use an LLM to filter the candidates and keep only content keywords, such as named entities and domain-specific terms.

\textit{(ii) Keyframe selection and generation.}
Given the reference text, the mined keyword, and the candidate keyframes, a vision language model (VLM) selects the frame that provides the strongest visual evidence for the keyword. In some cases, none of the sampled frames contains information relevant to the keyword; for these cases, we synthesize a substitute keyframe using GPT Image~2.0~\cite{openai2026chatgptimages2}. Synthesized frames account for about $10\%$ of the final benchmark.

\textit{(iii) Evidence classification.}
Given the keyword and its selected keyframe, the VLM further determines in what way the keyframe supports the keyword, and assigns one of the three evidence levels (ETE, EVE, and ISE).
Qwen3.5-35B-A3B~\cite{qwen2026qwen35vlm} is employed as the VLM in both (ii) and (iii).

\noindent\textbf{Stage 3: Expert Verification.}
As the annotations produced by the automatic stages are not always reliable, all remaining cases are manually verified by human annotators.
Each case is independently reviewed along four aspects: the
accuracy of the reference transcription, the difficulty of recognizing the keyword from audio alone, the visual relevance of the selected keyframe, and the correctness of the assigned
evidence level. A case is retained only if the keyword is relatively difficult to recognize from the audio alone, and the keyframe provides clear visual evidence for it. Finally, each retained case consists of an audio segment, a keyframe, a keyword, an evidence level, and a verified reference transcription.

\subsection{Visual Evidence Levels}
\label{sec:taxonomy}

We categorize the cases in \benchmark\ into three levels according to the relationship between the keyword and the visual evidence, with examples shown in Figure~\ref{fig:taxonomy}. From Level 1 to Level 3, the connection between the keyword and its
visual evidence becomes progressively less direct, which we call increasing semantic distance. This division allows us to examine how model performance changes as the visual evidence becomes harder to exploit.


\noindent\textbf{Level 1 -- Explicit Textual Evidence (ETE).}
The keyword appears directly as readable text in the frame, so it can be obtained by reading the text. In Figure~\ref{fig:taxonomy}(a), the word ``capitol'' is printed on the map, which distinguishes it from the acoustically similar ``capital.''

\noindent\textbf{Level 2 -- Explicit Visual Evidence (EVE).}
The keyword is no longer written in the frame, but the entity, object or action it refers to is directly shown. Identifying the keyword therefore requires recognizing the visual content and linking it to the spoken context, rather than simply reading text. In
Figure~\ref{fig:taxonomy}(b), the open landscape seen from the rooftop supports ``plain'' rather than ``plane''.

\noindent\textbf{Level 3 -- Implicit Semantic Evidence (ISE).}
The frame contains neither the keyword nor anything that directly corresponds to it, and the keyword can only be inferred from the scene as a whole, the domain it belongs to, or the related semantic context. This level therefore requires scene understanding and cross-modal reasoning.
In Figure~\ref{fig:taxonomy}(c), the frame shows a League of Legends match, and recognizing this domain favors the in-game ability name over the acoustically similar alternative.

\vspace{-0.3em}
\subsection{Dataset analysis}

\begin{figure}[t]
    \centering
    \includegraphics[scale=0.23]{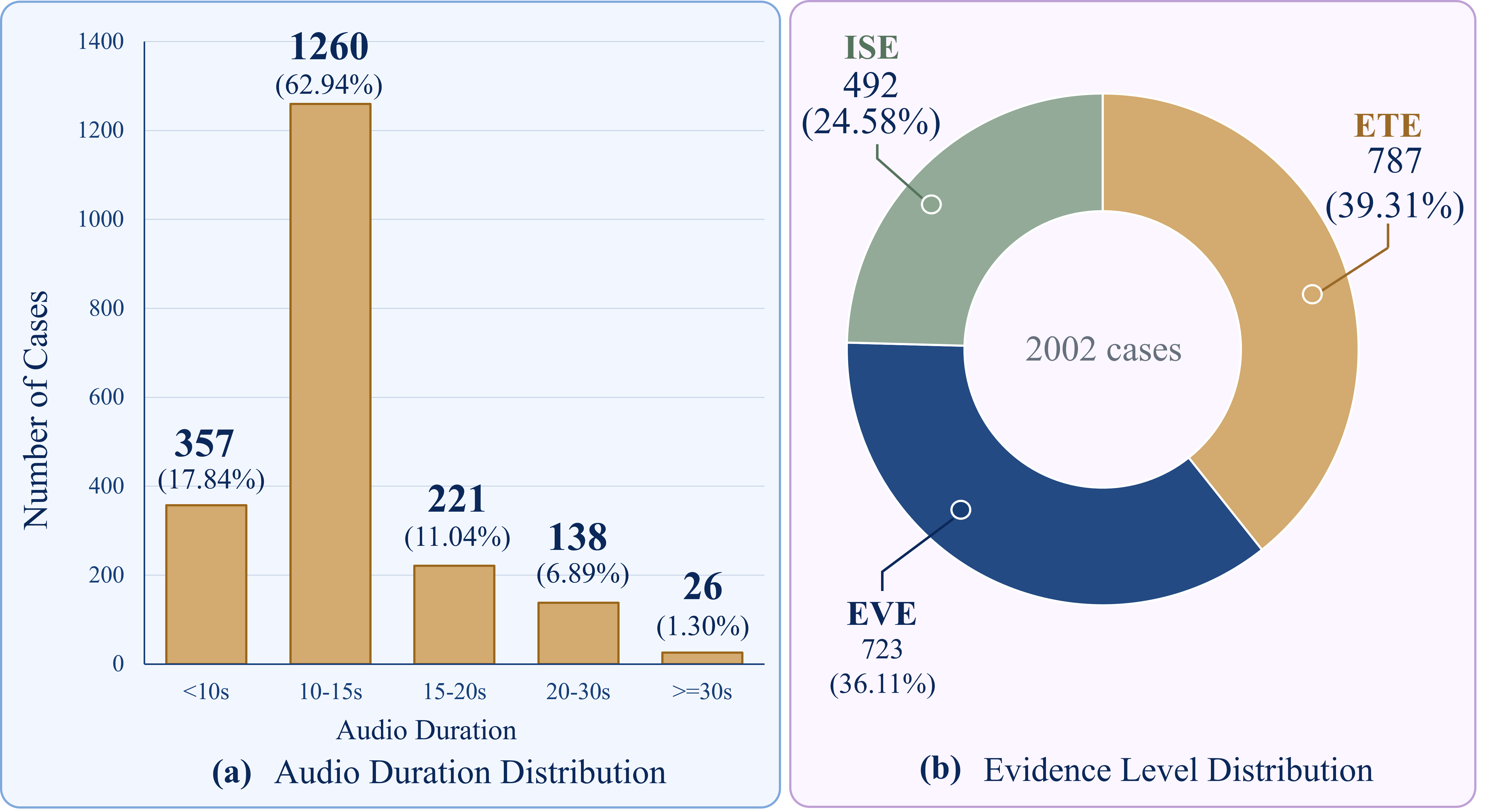}
    \caption{Statistics of the \benchmark.}
    \label{fig:dataset_analysis}
\end{figure}

Figure~\ref{fig:dataset_analysis} summarizes the statistics of \benchmark, which contains 2,002 cases. Most audio segments are short, with 91.82\% shorter than 20 seconds, a range comparable to that commonly used in ASR evaluation. For the evidence levels, ETE, EVE, and ISE account for 39.31\%, 36.11\%, and 24.58\% of the cases, respectively. ISE is the least frequent, as implicit semantic evidence is harder to collect than evidence explicitly presented in the keyframe, but all three levels still receive substantial coverage.

\section{Experiments}
\label{sec:experiments}

\subsection{Experimental Setup}

\begingroup
\setlength{\emergencystretch}{2em}
Experiments cover three model families: ASR systems, large audio language models (LALMs), and Omni models. The ASR systems are FireRedASR2-LLM~\cite{xu2026fireredasr2s}, MiMo-V2.5-ASR~\cite{xiaomi2026mimov25}, Qwen3-ASR-1.7B~\cite{shi2026qwen3asr}, Qwen-Audio-3.0-ASR~\cite{alibaba2026qwenaudio}, and Seed-ASR-2.0~\cite{bytedance2026seedasr}. The LALMs are MOSS-Audio-8B-Instruct~\cite{yang2026mossaudio}, MiMo-Audio-7B-Instruct~\cite{xiaomi2025mimoaudio}, and StepAudio-2.5-Chat~\cite{lin2026stepaudio25}. The Omni models are Qwen3-Omni~\cite{xu2025qwen3omni}, MiMo-V2.5~\cite{xiaomi2026mimov25}, Qwen3.5-Omni-Plus~\cite{qwen2026qwen35omni}, and Doubao-Seed-2.0-lite~\cite{bytedance2026seed20}.

All models are evaluated with speech alone, and models that support context are further evaluated with visual context in two forms: a textual caption of the keyframe generated by Qwen3.5-35B-A3B~\cite{qwen2026qwen35vlm}, and the keyframe image itself. ASR systems and LALMs take only the caption, whereas omni models take both.

\par
\endgroup

\subsection{Evaluation Metrics}
\label{sec:metrics}

System performance is evaluated using WER, Recall, and normalized visual gain (NVG). WER is used to measure overall transcription accuracy, whereas Recall is used to evaluate the recognition of annotated keywords:
\begin{equation}
    \mathrm{Recall}=\frac{N_{\mathrm{hit}}}{N_{\mathrm{total}}},
\end{equation}
where $N_{\mathrm{total}}$ is the total number of annotated keywords and $N_{\mathrm{hit}}$ denotes the number of keywords correctly recognized by the model.

To quantify the improvement brought by visual context, we further introduce NVG:
\begin{equation}
    \mathrm{NVG}=\frac{Recall_{\mathrm{A+V}}-Recall_{\mathrm{A}}}{1-Recall_{\mathrm{A}}},
\end{equation}
where $Recall_{\mathrm{A}}$ and $Recall_{\mathrm{A+V}}$ denote recall under the audio-only and audio-visual conditions, respectively. An NVG of 0 indicates no improvement over the audio-only baseline, whereas an NVG of 1 indicates that perfect recall is achieved under the audio-visual condition. Negative values indicate that keyword recognition is degraded by the incorporation of visual context.

\subsection{Main Results}

\begin{table}[t]
\centering
\caption{Level-wise visual-context utilization. Recall, $\Delta$Recall, and NVG are averaged within each family model; $\Delta$Recall denotes the absolute recall gain.
}
\label{tab:hierarchy}

\small
\setlength{\tabcolsep}{2.55pt}
\renewcommand{\arraystretch}{1.05}

\begin{tabular}{lccccc}
\toprule
\multirow{2}{*}[-0.7ex]{\textbf{Family}} &
\multirow{2}{*}[-0.7ex]{\textbf{Level}} &
\multicolumn{2}{c}{\textbf{Recall (\%)}} &
\multirow{2}{*}[-0.7ex]{\textbf{$\Delta$Recall}} &
\multirow{2}{*}[-0.7ex]{\textbf{NVG(\%)}} \\
\cmidrule(lr){3-4}
& & \textbf{Audio-only} & \textbf{Audio + Vision} & & \\
\midrule

\multirow{3}{*}{ASR}
& ETE & 42.64 & 69.17 & 26.53 & 45.16 \\
& EVE & 54.40 & 61.35 & 6.95  & 15.13 \\
& ISE & 47.90 & 52.29 & 4.39  & 8.33  \\

\midrule

\multirow{3}{*}{LALM}
& ETE & 43.21 & 78.24 & 35.03 & 61.34 \\
& EVE & 51.85 & 58.31 & 6.46  & 13.37 \\
& ISE & 46.91 & 50.92 & 4.01  & 7.49  \\

\midrule

\multirow{3}{*}{Omni}
& ETE & 58.72 & 84.38 & 25.66 & 62.24 \\
& EVE & 65.79 & 75.18 & 9.39  & 27.08 \\
& ISE & 61.61 & 68.87 & 7.26  & 18.88 \\

\bottomrule
\end{tabular}
\end{table}

Table~\ref{tab:model_results} reports experimental results on three model families and Table~\ref{tab:hierarchy} summarizes visual-context utilization across three evidence levels. Several general observations can be drawn from these results.

\noindent\textbf{Beyond WER: Keyword Recall.} Under the audio-only condition, Seed-ASR-2.0 achieves the lowest WER of 4.82\% but only 54.09\% recall. In contrast, omni models generally achieve higher recall, with Qwen3.5-Omni-Plus reaching 65.83\%. These results show that lower WER does not necessarily indicate better recognition of target keywords. Since this benchmark focuses on target keywords, recall directly measures whether they are correctly recognized.

\noindent\textbf{Benefits of visual context.} Across all three model families, incorporating visual context consistently improves keyword recognition. Specifically, average recall increases from 48.18\% to 62.20\% for ASR models, from 47.24\% to 64.33\% for LALMs, and from 61.98\% to 77.24\% for omni models. At the same time, average WER decreases from 6.02\% to 5.19\%, from 8.19\% to 6.65\%, and from 6.47\% to 4.74\%, respectively. These results demonstrate that visual context provides effective cues for recognizing challenging keywords while generally improving overall transcription accuracy.

\noindent\textbf{Effects of visual input representation.}
The effectiveness of visual context may depend on how the visual information is represented. For Qwen3-Omni, caption and image inputs achieve comparable overall recall (74.09\% vs.\ 74.89\%), while caption input yields a substantially lower WER (4.01\% vs.\ 5.49\%). The other omni models exhibit a similar pattern. This suggests that textualized visual information can be more effectively used for transcription. A possible explanation is that captions provide compact information closely aligned with the speech content, whereas raw images may introduce irrelevant visual information. More effective use of raw visual context therefore remains an important direction for omni models.

\noindent\textbf{Visual gains decrease with semantic distance.} 
As shown in Table~\ref{tab:hierarchy}, visual context improves recall across all three model families, but the magnitude of the improvement varies substantially across evidence levels. The absolute recall gain is consistently highest for ETE, followed by EVE and ISE. For ASR models, the average recall gain decreases from 26.53\% for ETE to 6.95\% for EVE and 4.39\% for ISE, with the same pattern observed for LALMs and omni models. Mean NVG follows a similar level-wise trend across all three families, confirming that both absolute and normalized gains diminish as the semantic distance between the visual evidence and the target term increases. These results indicate that visual context is more effective when the evidence is explicit and directly related to the keyword, while current models struggle to exploit implicit visual cues, highlighting the need to improve their ability to reason over and integrate such evidence.

\section{Conclusion}
\label{sec:conclusion}

We introduce \benchmark, a human-verified, tiered benchmark for evaluating how effectively speech models leverage visual context to recognize challenging keywords. It covers three levels of visual evidence: ETE, EVE, and ISE. Across ASR models, LALMs, and omni models, visual context consistently improves keyword recall. However, the larger gains on ETE than on EVE and ISE show that current models exploit on-screen text more effectively than indirect visual evidence, highlighting the need for stronger visual grounding and reasoning. We hope \benchmark\ will provide a useful foundation for future research on multimodal ASR.

\clearpage
\balance
\bibliographystyle{IEEEbib}
\bibliography{refs}

@inproceedings{panayotov2015librispeech,
  author = {Panayotov, Vassil and Chen, Guoguo and Povey, Daniel and Khudanpur, Sanjeev},
  title     = {{LibriSpeech}: An {ASR} Corpus Based on Public Domain Audio Books},
  booktitle = {Proc. ICASSP},
  pages={5206--5210},
  year      = {2015}
}

@inproceedings{radford2023whisper,
  author={Radford, Alec and Kim, Jong Wook and Xu, Tao and Brockman, Greg and McLeavey, Christine and Sutskever, Ilya},
  title     = {Robust Speech Recognition via Large-Scale Weak Supervision},
  booktitle = {Proc. ICML},
  pages={28492--28518},
  year      = {2023}
}

@inproceedings{gabeur2022avatar,
  author={Gabeur, Valentin and Seo, Paul Hongsuck and Nagrani, Arsha and Sun, Chen and Alahari, Karteek and Schmid, Cordelia},
  title     = {{AVATAR}: Unconstrained Audiovisual Speech Recognition},
  booktitle = {Proc. Interspeech},
  pages     = {2818--2822},
  year      = {2022}
}

@inproceedings{wang2024slidespeech,
  author = {Wang, Haoxu and Yu, Fan and Shi, Xian and Wang, Yuezhang and Zhang, Shiliang and Li, Ming},
  title     = {{SlideSpeech}: A Large-Scale Slide-Enriched Audio-Visual Corpus},
  booktitle = {Proc. ICASSP},
  pages={11076--11080},
  year      = {2024}
}

@inproceedings{wang2024slideavsr,
  author={Wang, Hao and Kurita, Shuhei and Shimizu, Shuichiro and Kawahara, Daisuke},
  title     = {{SlideAVSR}: A Dataset of Paper Explanation Videos for Audio-Visual Speech Recognition},
  booktitle = {Proc. ALVR},
  pages={129--137},
  year      = {2024}
}

@inproceedings{sinhamahapatra2025slides,
  author={Sinhamahapatra, Supriti and Niehues, Jan},
  title     = {Do Slides Help? Multi-modal Context for Automatic Transcription of Conference Talks},
  booktitle = {Proc. EMNLP},
  pages={16111--16121},
  year      = {2025}
}

@inproceedings{hu2026vapo,
  author={Hu, Rui and Qiu, Delai and Wang, Yining and Liu, Shengping and Sang, Jitao},
  title     = {{VAPO}: End-to-End Slide-Enhanced Speech Recognition with Omni-Modal Large Language Models},
  booktitle = {Proc. ACL},
  pages = {9417--9432},
  year      = {2026}
}

@inproceedings{zhao2025chineselips,
  author={Zhao, Jinghua and Jia, Yuhang and Wang, Shiyao and Zhou, Jiaming and Wang, Hui and Qin, Yong},
  title     = {{Chinese-LiPS}: A Chinese Audio-Visual Speech Recognition Dataset with Lip-Reading and Presentation Slides},
  booktitle = {Proc. ICME},
  pages={1--6},
  year      = {2025}
}

@article{xu2026fireredasr2s,
  author={Xu, Kaituo and Jia, Yan and Huang, Kai and Chen, Junjie and Li, Wenpeng and Liu, Kun and others},
  title   = {{FireRedASR2S}: A State-of-the-Art Industrial-Grade All-in-One Automatic Speech Recognition System},
  journal = {arXiv preprint arXiv:2603.10420},
  year    = {2026}
}

@article{shi2026qwen3asr,
  author={Shi, Xian and Wang, Xiong and Guo, Zhifang and Wang, Yongqi and Zhang, Pei and Zhang, Xinyu and others},
  title   = {{Qwen3-ASR} Technical Report},
  journal = {arXiv preprint arXiv:2601.21337},
  year    = {2026}
}

@article{alibaba2026qwenaudio,
  author={Bian, Chuanmeng and Chen, Daren and Chen, Peixin and Chen, Zhigao and Fan, Zhiyun and Gao, Zhifu and others},
  title   = {{Qwen-Audio-3.0-ASR} Technical Report},
  journal = {arXiv preprint arXiv:2609.07549},
  year    = {2026}
}

@article{yang2026mossaudio,
  author={Yang, Chen and Yu, Chufan and Chen, Hanfu and Zhu, Jie and Chen, Jingqi and Chen, Ke and others},
  title   = {{MOSS-Audio} Technical Report},
  journal = {arXiv preprint arXiv:2606.01802},
  year    = {2026}
}

@article{xiaomi2025mimoaudio,
  author = {{Xiaomi LLM-Core Team}},
  title   = {{MiMo-Audio}: Audio Language Models Are Few-Shot Learners},
  journal = {arXiv preprint arXiv:2512.23808},
  year    = {2025}
}

@misc{bytedance2026seedasr,
  author       = {{ByteDance Seed / Volcano Engine}},
  title        = {{Seed-ASR 2.0}},
  howpublished = {Official model documentation},
  year         = {2026}
}

@article{xu2025qwen3omni,
  author={Xu, Jin and Guo, Zhifang and Hu, Hangrui and Chu, Yunfei and Wang, Xiong and He, Jinzheng and others},
  title   = {{Qwen3-Omni} Technical Report},
  journal = {arXiv preprint arXiv:2509.17765},
  year    = {2025}
}

@misc{xiaomi2026mimov25,
  author       = {{Xiaomi MiMo Team}},
  title        = {{MiMo-V2.5}},
  howpublished = {Official model card/documentation},
  year         = {2026}
}

@misc{qwen2026qwen35vlm,
  author       = {{Qwen Team}},
  title        = {{Qwen3.5}: Towards Native Multimodal Agents},
  howpublished = {Official model documentation},
  year         = {2026},
  url          = {https://qwen.ai/blog?id=qwen3.5}
}

@article{qwen2026qwen35omni,
  author  = {{Qwen Team}},
  title   = {{Qwen3.5-Omni} Technical Report},
  journal = {arXiv preprint arXiv:2604.15804},
  year    = {2026}
}

@article{bytedance2026seed20,
  author  = {{ByteDance Seed}},
  title   = {{Seed2.0} Model Card: Towards Intelligence Frontier for Real-World Complexity},
  journal = {arXiv preprint arXiv:2607.00248},
  year    = {2026}
}

@inproceedings{ghorbani2021listen,
  author={Ghorbani, Shahram and Gaur, Yashesh and Shi, Yu and Li, Jinyu},
  title     = {Listen, Look and Deliberate: Visual Context-Aware Speech Recognition Using Pre-Trained Text-Video Representations},
  booktitle = {Proc. SLT},
  pages={621--628},
  year      = {2021}
}

@article{tian2026seeing,
  author={Tian, Wenjie and Shao, Mingchen and Mu, Bingshen and Geng, Xuelong and Wang, Chengyou and Liao, Yujie and others},
  title   = {Seeing the Context: Rich Visual Context-Aware Speech Recognition via Multimodal Reasoning},
  journal = {arXiv preprint arXiv:2603.07263},
  year    = {2026}
}

@inproceedings{chen2026ocr,
  author={Chen, Junli and Tang, Changli and Li, Yixuan and Sun, Guangzhi and Zhang, Chao},
  title     = {{OCR}-Enhanced Multimodal {ASR} Can Read While Listening},
  booktitle = {Proc. ICASSP},
  pages={18482--18486},
  year      = {2026}
}

@article{lin2026stepaudio25,
  title={StepAudio 2.5 Technical Report},
  author={Lin, Bin and Zhao, Bo and Wu, Boyong and Yan, Chao and Wu, Chen and Yi, Cheng and others},
  journal={arXiv preprint arXiv:2605.23463},
  year={2026}
}

@inproceedings{aishell1,
  title     = {{AISHELL-1}: An open-source mandarin speech corpus and a speech recognition baseline},
  author={Bu, Hui and Du, Jiayu and Na, Xingyu and Wu, Bengu and Zheng, Hao},
  booktitle = {Proc. Oriental COCOSDA},
  pages={1--5},
  year      = {2017}
}

@article{aishell2,
  author={Du, Jiayu and Na, Xingyu and Liu, Xuechen and Bu, Hui},
  title   = {{AISHELL-2}: Transforming Mandarin {ASR} Research Into Industrial Scale},
  journal = {arXiv preprint arXiv:1808.10583},
  year    = {2018}
}

@inproceedings{salmonn,
  author={Tang, Changli and Yu, Wenyi and Sun, Guangzhi and Chen, Xianzhao and Tan, Tian and Li, Wei and others},
  title     = {{SALMONN}: Towards Generic Hearing Abilities for Large Language Models},
  booktitle = {Proc. ICLR},
  pages={16607--16629},
  year      = {2024}
}

@article{slam,
  title={{SLAM-LLM}: A Modular, Open-Source Multimodal Large Language Model Framework and Best Practice for Speech, Language, Audio and Music Processing},
  author={Ma, Ziyang and Yang, Guanrou and Chen, Wenxi and Gao, Zhifu and Du, Yexing and Li, Xiquan and others},
  journal={IEEE Journal of Selected Topics in Signal Processing},
  year={2026}
}

@article{qwen2-audio,
  title = {{Qwen2-Audio} Technical Report},
  author={Chu, Yunfei and Xu, Jin and Yang, Qian and Wei, Haojie and Wei, Xipin and Guo, Zhifang and others},
  journal={arXiv preprint arXiv:2407.10759},
  year={2024}
}

@article{index-asr,
  title={{Index-ASR} technical report},
  author={Song, Zheshu and Wang, Lu and Deng, Wei and Yang, Zhuo and Wu, Yong and Xia, Bin},
  journal={arXiv preprint arXiv:2601.00890},
  year={2025}
}

@article{review,
  title={Recent advances in end-to-end automatic speech recognition},
  author={Li, Jinyu},
  journal={arXiv preprint arXiv:2111.01690},
  year={2021}
}

@misc{openai2026chatgptimages2,
  author = {{OpenAI}},
  title  = {{System Card: ChatGPT Images 2.0 and Thinking Mode}},
  year   = {2026},
  month  = apr
}

@inproceedings{MLS,
  author={Vineel Pratap and Qiantong Xu and Anuroop Sriram and Gabriel Synnaeve and others},
  title={{MLS}: A Large-Scale Multilingual Dataset for Speech Research},
  booktitle={Proc. Interspeech},
  pages     = {2757--2761},
  year={2020},
}

@article{funaudiollm,
  title         = {{FunAudioLLM}: Voice Understanding and Generation Foundation Models for Natural Interaction Between Humans and {LLMs}},
  author        = {An, Keyu and Chen, Qian and Deng, Chong and Du, Zhihao and Gao, Changfeng and Gao, Zhifu and others},
  year          = {2024},
  journal  =  {arXiv preprint arXiv:2407.04051}
}

@inproceedings{fleurs,
  author={Conneau, Alexis and Ma, Min and Khanuja, Simran and Zhang, Yu and Axelrod, Vera and Dalmia, Siddharth and others},
  booktitle={Proc. SLT}, 
  title={{FLEURS}: FEW-Shot Learning Evaluation of Universal Representations of Speech}, 
  year={2023},
  pages={798--805}
}

\end{document}